\documentclass[sigconf]{acmart}
\AtBeginDocument{%
  }

\setcopyright{none}              %
\renewcommand\footnotetextcopyrightpermission[1]{} %

\hypersetup{
    colorlinks = true,
    linkcolor = blue,
    anchorcolor = blue,
    citecolor = blue,
    filecolor = blue,
    urlcolor = blue
}

\newcommand{\ignore}[1]{}

\begin{document}

\title{The Natural Language Interaction Protocol and Standard \\for AI Agents 
}

\author{%
Luyi Xing\textsuperscript{1},
Rasit Onur Topaloglu\textsuperscript{2},
Ranjan Sinha\textsuperscript{3},
Abhay Ratnaparkhi\textsuperscript{4},
Samuel Ndichu\textsuperscript{5},
Christopher Nguyen\textsuperscript{6},
Anindita Das\textsuperscript{9},
Tom Sheffler\textsuperscript{7},
Mohamed Rahouti\textsuperscript{8},
Zichuan Li\textsuperscript{1},\\
Xiaojing Liao\textsuperscript{1},
Sanjay Aiyagari\textsuperscript{9}
}

\affiliation{%
  \institution{%
  \textsuperscript{1}University of Illinois at Urbana-Champaign;
  \textsuperscript{2}Marist University;
  \textsuperscript{3}IBM;
  \textsuperscript{4}eBay;
  \textsuperscript{5}NICT, Japan;
  \textsuperscript{6}Aitomatic, Inc.;
  \textsuperscript{7}Lenovo;
  \textsuperscript{8}Fordham University;
  \textsuperscript{9}Red Hat
  }
}

\email{%
\{lxing2,zichuan7,xjliao\}@illinois.edu,
rasit.topaloglu@marist.edu,
anidas@redhat.com,
rsinha@us.ibm.com,
abratnaparkhi@ebay.com,
ndichu@nict.go.jp,
ctn@aitomatic.com,
tsheffler@lenovo.com,
mrahouti@fordham.edu,
sanjay@redhat.com
}

\renewcommand{\shortauthors}{Xing et al.}

\begin{abstract}
AI agents are increasingly being developed and deployed across organizations using heterogeneous agent-development frameworks, AI models, tool interfaces, protocols, and execution environments. To realize their potential social and business impact, these agents must be able to interoperate through a common communication protocol. The Natural Language Interaction Protocol (NLIP), developed by researchers and practitioners across companies and universities and standardized by Ecma International, addresses this need by defining a standards-based application-layer protocol for AI-agent interaction. NLIP provides a lightweight semantic message envelope that can be carried over existing transports such as HTTP/HTTPS, WebSocket, and AMQP, while allowing NLIP-aware agents and gateways to adapt between clients, agents, local context stores, ontologies, tools, enterprise services, and heterogeneous underlying protocols. This paper presents the motivation and design rationale of NLIP, its message model and transport bindings, security-by-design considerations, reference implementation, representative applications, adoption signals, and relationship to emerging agent protocols such as MCP and A2A.
\end{abstract}

\maketitle
\pagestyle{plain}

\section{Introduction}

AI-based software systems, or agents, are beginning to be deployed in a wide range of industrial applications. To realize their full potential, agents serving different functionalities and operated by different organizations must interoperate through a common communication protocol. Motivated by this need, a group of academic and industrial researchers collaborated to develop the \emph{Natural Language Interaction Protocol (NLIP)}. NLIP was formally standardized in December 2025 by Ecma International Technical Committee 56 (TC56) as ECMA-430, together with related transport bindings and security-profile specifications~\cite{1_ECMA430,2_ECMA431,3_ECMA432,4_ECMA433,5_ECMA434}. In addition to Ecma International, the work leading to the standard has been supported by the Enterprise Neurosystem Group (enterpriseneurosystem.org) and the AI Alliance (thealliance.ai), two open-source organizations promoting collaboration across the broader AI community. This paper introduces the NLIP protocol, its key design innovations, security considerations, reference implementation, adoption signals, representative applications, and relation with relevant protocols. \looseness=-1

\begin{figure}
  \includegraphics[width=1.0\linewidth ]{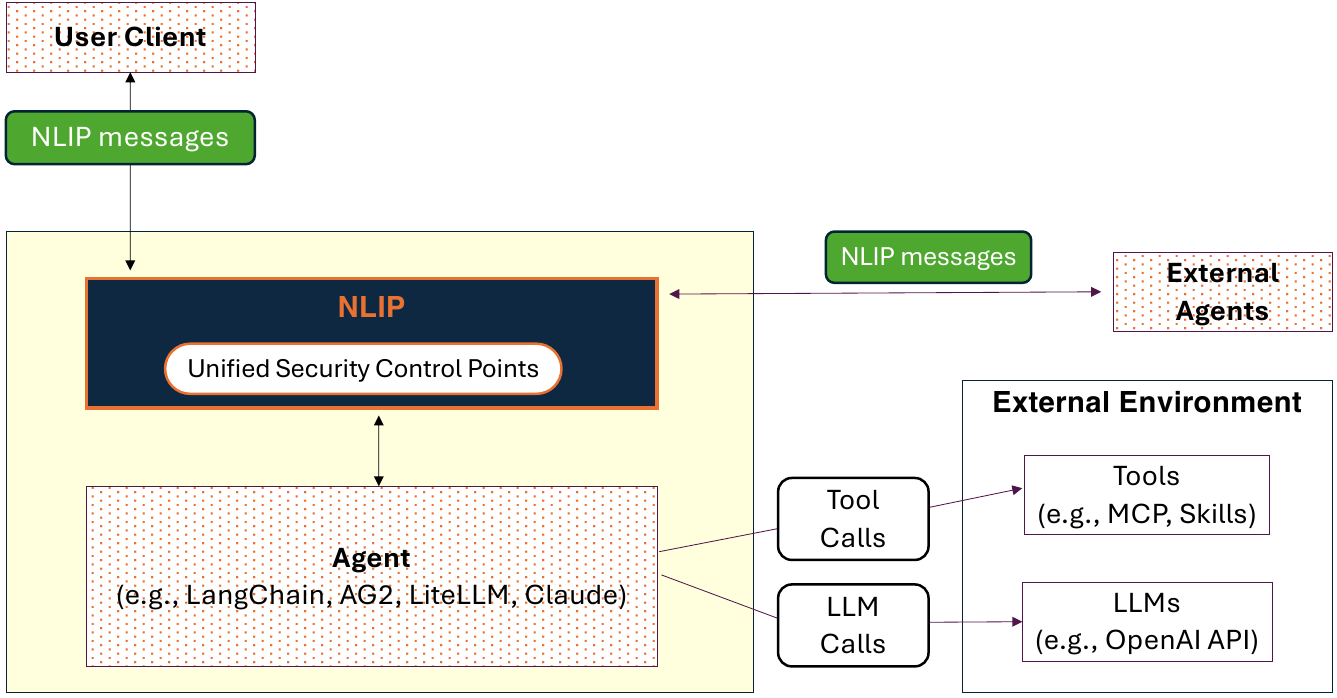}
  
  \caption{NLIP as an application-layer interoperability layer connecting user clients, agents, tools, LLMs, and external agents through standardized NLIP messages.}
  \label{fig:nlip_architecture}
  \vspace{-15pt}
\end{figure}

Rather than replacing existing agent frameworks, tool protocols, or enterprise services, NLIP defines a standardized interaction layer that is independent of agent-internal technologies, logic, and AI models being used by the agents. NLIP is an application-layer protocol: it standardizes agent interaction semantics and message envelopes while directly leveraging existing underlying transport protocols, including HTTP/HTTPS, WebSocket, and AMQP. NLIP-aware agents and gateways expose a common interface to user clients and other agents while continuing to use their native frameworks, LLM APIs, tool protocols, and enterprise services internally. As outlined in Figure~\ref{fig:nlip_architecture}, lightweight NLIP adapters enable heterogeneous agents to communicate through a common protocol without modifying their internal reasoning or execution logic.
This architecture positions NLIP as a standards-based interoperability layer for heterogeneous agent ecosystems, enabling independently developed agents to communicate while preserving existing investments in frameworks, tools, and enterprise infrastructure.

In the following, \S~\ref{sec_origins} describes origins of NLIP including when the idea was introduced and started to be developed.
\S~\ref{sec_innovation} describes the core design innovation of NLIP. \S~\ref{sec_design_overview} provides an overview of the NLIP protocol design, including its message model and transport bindings. \S~\ref{sec_security_by_design} discusses security by design for agent communication. \S~\ref{sec_reference_impl} presents the reference implementation, development framework adoption, and representative applications. \S~\ref{sec_other_protocol} positions NLIP within the broader landscape of agent protocols and reports an empirical comparison with A2A (\S~\ref{sec_comparision_with_a2a}). \S~\ref{sec_conclusion} concludes. The Appendix provides additional details on the origins and standardization history of NLIP. \looseness=-1

\section{Origins of NLIP}
\label{sec_origins}

NLIP started in June 2024 as a technical working group within the Enterprise NeuroSystem Group (ENG). ENG is an open-source consortium involving 30+ companies and universities formed with the goal of enabling technologies needed to establish a global AI network. The original members of the working group included researchers from the University of Delaware, the University at Buffalo, the University of Michigan, SRI International, IBM, and Red Hat. The group eventually expanded to further include researchers from Cisco, Fordham University, Pennsylvania State University, and Indiana University Bloomington. In October 2024, the initial version of the protocol was defined, and a decision to convert the specification to a formal standard was made. Ecma International, the group that defines various application-level technologies such as JavaScript, was identified as the one closest to the needs of an application-level protocol design, and the group was formally inducted as a technical working group within ECMA in December 2024. The effort was joined by ServiceNow and Hitachi, which supported the formation of the group in ECMA.  In early 2025, ENG became an official member of the AI Alliance, a consortium of several companies and universities involved in promoting open-source AI technologies and governance. The Alliance decided to incorporate the working group as part of its activities, providing resources such as a GitHub space for the group to continue its activities as well. Subsequently, researchers from several other organizations joined the effort, including those from Qualcomm, Japan National Institute of Information and Communications Technology (NICT), University of Illinois Urbana-Champaign (UIUC), Marist University, Chicago State University, City University of New York, Microsoft, and Palo Alto Networks. 

Ecma requires formal membership of organizations in order to participate, while ENG allows open participation of any technical person from any organization. The members who officially joined Ecma TC56 to define the official standards included IBM, Hitachi, Red Hat, ServiceNow, Purdue University, Indiana University Bloomington, University at Buffalo, Fordham University, Chicago State University, Marist University, and City University of New York. Researchers at other organizations contributed only as technical advisors as part of the ENG working group. The ENG group met weekly to discuss technical issues and implement prototypes, while the Ecma standard team met monthly to formalize the standards. This bi-pronged approach allowed us to make rapid technical progress while making steady progress towards the stage of official standardization. 

The ENG effort also allowed the group to explore the design of NLIP through several proof-of-concept prototypes, which are described in the NLIP explanatory report and summarized later in this paper~\cite{6_ECMATR113}. 

\section{Design Innovations of NLIP}
\label{sec_innovation}

Traditional application-layer protocols typically define fixed structures exchanged between communicating parties, such as JSON messages, web forms, or XML documents. This approach works well when clients and servers share a stable data model, but it becomes brittle in heterogeneous agent ecosystems, where agents, tools, models, ontologies, and task representations may evolve independently. Even modest changes to one endpoint's internal representation can break communication unless schemas, versions, and compatibility rules are carefully managed. This tight coupling is a fundamental source of API-management complexity in distributed service environments and becomes especially limiting for AI-agent systems that must interoperate across organizations, frameworks, and application domains.

\begin{figure}
  \includegraphics[width=0.8\linewidth ]{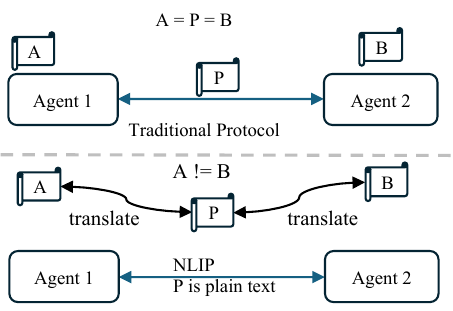}
  
  \caption{The distinctive design of NLIP}
  \label{fig:nlip_abstract}
  \vspace{-10pt}
\end{figure}

NLIP takes a different approach, which decouples the internal structure representation between the client and the server. Instead of specifying a rigid structure on the wire, natural language is used to transport the semantics of the information that the client and server need to exchange. Each side, the client and the server, can use an internal AI model to translate the natural language to its internal representation. This allows the client and server to maintain a different internal representation, if needed~\cite{1_ECMA430}. Because modern AI models are increasingly effective at translating between structured and unstructured versions of information, NLIP allows the definition of an application-level protocol that is more flexible and easier to maintain. This fundamental difference is shown in Figure~\ref{fig:nlip_abstract}. \looseness=-1

In practice, this translation can be performed by NLIP-aware agents or gateways. These endpoints form the adaptive layer: they receive NLIP messages, ground them in local context, and translate them into the internal representations, tools, APIs, ontologies, or protocols used by the receiving system. 
Because NLIP relies on semantic translation rather than rigid shared schemas, implementations should guard against semantic drift, including context loss or misinterpretation, when domain-specific meanings cross protocol boundaries. NLIP-aware agents and gateways can mitigate this by grounding NLIP messages in structured submessages, message-type hints, labels, provenance, ontology or entity references, and local validation rules where appropriate. In this sense, NLIP does not replace schemas, ontologies, or domain models; it provides a semantic envelope through which they can be referenced, mediated, and preserved across heterogeneous systems.

When the client and server need to exchange modalities different from text, e.g., video or audio, a similar paradigm can be followed. NLIP provides an envelope protocol that describes the type of content being exchanged, text, audio, video, or another modality, and lets the endpoints translate it to their internal representation.

\section{NLIP Design Overview}
\label{sec_design_overview}

\begin{figure}
  \includegraphics[width=0.65\linewidth ]{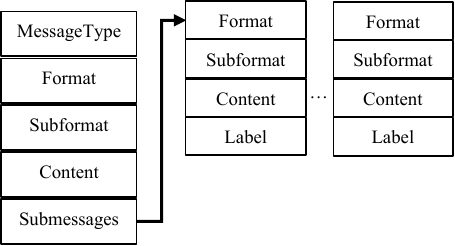}
  
  \caption{Structure of an NLIP message}
  \label{fig:nlip_msg}
  \vspace{-20pt}
\end{figure}

\noindent\textbf{Message model.}
To enable a multi-modal, flexible interaction among two parties involved in a communication, NLIP exchanges messages as JSON objects~\cite{1_ECMA430}. The message consists primarily of five fields, of which three are required, and two are optional. The three required fields include a content field, which contains the actual information being exchanged, a format field, which describes the format of the content, and a subformat field, which defines a further refinement of the format field. As an example, the content exchanged may be general text, the format would specify it as text, and the subformat would specify the language of the content, e.g., English. The optional fields include a message-type designation and an array of submessages. The message-type field can contain an application-specific string that can help the other side in parsing the content, while the submessage field allows several other pieces of content to be carried in the same message. The structure is shown in Figure~\ref{fig:nlip_msg}. \looseness=-1

Each entry in the array of submessages contains the three required fields of format, subformat, and content, and it contains an optional label field. The label, like the message type, can be used as hints to allow communicating parties to define their own conventions for rapid processing of content. As an example, the label field may indicate whether a submessage identifying location information is the pick-up location or the drop-off location for a journey. \looseness=-1

The NLIP envelope is intentionally lightweight: it standardizes the interaction boundary without prescribing the internal ontology, reasoning method, model, or execution framework of each endpoint. This allows NLIP-aware agents to carry natural-language intent, structured context, provenance, constraints, and references to domain entities while preserving local implementation choices. \looseness=-1

The format field may identify that the content being exchanged is natural text, with the subformat identifying the language. This is expected to be the predominant mode of information exchange. Other values of format fields include identifying a content as token with the subformat identifying whether it is a correlation token (allows multiple related messages in a communication stream to be correlated) or an authentication token (to allow for authentication of communicating parties); indicating that the content is a structured text with the subformat identifying whether it is XML, JSON, HTML or something else; indicating that the content is binary data with the subformat identifying whether it is an audio with a specific encoding, video data with a specific encoding, sensor data etc.; marking a content as a location with the subformat identifying if the location is textual description or latitude/longitude description; and a generic format which can allow for arbitrary information to be processed by the other side. 

\vspace{3pt}\noindent\textbf{Transport bindings.}
As an application-layer protocol, NLIP does not define a new network transport. Instead, it defines a lightweight message envelope and interaction semantics that can be carried over existing transport protocols. The current NLIP standards define official bindings over HTTP/HTTPS~\cite{2_ECMA431}, WebSocket~\cite{3_ECMA432}, and AMQP~\cite{4_ECMA433}. This design allows NLIP endpoints to reuse widely deployed transport infrastructure while providing a common semantic interaction layer for clients, agents, and gateways.

\section{Security by Design for Agent Communication}
\label{sec_security_by_design}

\vspace{3pt}\noindent\textbf{Security profiles.} NLIP defines security profiles for deployments with different assurance requirements~\cite{5_ECMA434}. However, AI-agent communication introduces risks beyond traditional transport security, including prompt injection, indirect prompt injection, unsafe tool invocation, sensitive-data leakage, and loss of provenance across agent chains~\cite{14_OWASPPromptInjection,15_IndirectPromptInjection,16_InjecAgent}. NLIP defines three types of security profiles that should be supported by NLIP-based systems~\cite{5_ECMA434}. 
These profiles range from basic security requirements to more rigorous enterprise-oriented requirements intended to minimize common security risks in NLIP-based deployments.
The rigorous security profile is designed to address the requirements of enterprise agents, which are often not addressed in commonly available standards or technical publications.

\noindent\textbf{Unified control points for AI and agent security.}
Beyond defining security profiles, NLIP's standardized communication layer provides a natural foundation for security enforcement across heterogeneous agent ecosystems. Because all communication between user clients and agents, and among agents themselves, is carried through standardized NLIP messages, validation, authorization, provenance tracking, policy enforcement, auditing, and security monitoring can be applied consistently at the communication layer rather than being fragmented across individual agent frameworks or applications. This provides a common foundation for mitigating both \emph{AI-driven vulnerabilities}---such as prompt injection, indirect prompt injection, chain-of-thought leakage, semantic manipulation, and privacy violations~\cite{14_OWASPPromptInjection,15_IndirectPromptInjection,16_InjecAgent,green-etal-2025-leaky,Zheng2025MISTJB,wang2025privacy,usenix-infer24}---and \emph{system-level vulnerabilities}---such as session hijacking, confused-deputy and token-passthrough attacks, unauthenticated inference flooding, memory injection attacks, and data-at-rest exposure~\cite{unit42_session_smuggling_2025,roychowdhury2024confusedpilot,mcp_security_best_practices_2025,kumar2025overthink,owasp_llm10_unbounded_consumption_2025,shafran2025machine,chen2024agentpoison,zou2025poisonedrag,cui2025safeguard}. By providing a unified enforcement point for agent communication (Figure~\ref{fig:nlip_architecture}), NLIP enables consistent protection across independently developed agent frameworks while preserving interoperability.

\section{Implementation and Example Applications}
\label{sec_reference_impl}

The flexibility and ease of use of NLIP have been explored through several proof-of-concept implementations at participating organizations. These examples illustrate a common architectural pattern: NLIP-aware agents act as adaptive interfaces between a standardized NLIP interaction layer and heterogeneous underlying systems, including tool protocols, enterprise APIs, knowledge bases, domain ontologies, industrial standards, and vendor-specific services. Several of these use cases have been described in the technical report~\cite{6_ECMATR113} describing NLIP motivation. Those include the use of NLIP as a question-answering service, as a proxy federating across several agents, as a gateway in front of a service using MCP to invoke tools~\cite{6_ECMATR113,10_MCPSpecification}, a speech-based customer-service agent, and as an integrator of services in a hybrid cloud environment (specifically integrating Active Directory with a ServiceNow System). In this pattern, NLIP does not replace MCP. Instead, an NLIP-aware agent or gateway exposes a standard NLIP interface to the client while using MCP internally to invoke tools and services. MCP remains the tool/service protocol, while NLIP provides the interaction envelope for agents and clients.

\subsection{Reference Implementation of NLIP}
The NLIP protocol is accompanied by open-source reference implementations and software development kits (SDKs), including \texttt{nlip\_sdk}, \texttt{nlip\_client}, and \texttt{nlip\_server}~\cite{21_NLIPProjectDocuments}. Together, these libraries implement the core NLIP protocol, including message structures, transport bindings, serialization, and client/server communication primitives. The SDKs are intentionally independent of agent-internal application logic and development frameworks, allowing existing agents and applications to adopt NLIP without modifying their internal execution models. This separation enables developers to focus on application logic while relying on reusable, standards-compliant implementations for agent communication, facilitating interoperability across heterogeneous AI-agent ecosystems. \looseness=-1

\subsection{Extension to Agent Development Frameworks and Adoption Signals}
The framework-independent architecture of NLIP enables existing agent systems to expose standardized NLIP interfaces through lightweight adapters, without changing their internal reasoning, planning, or tool-invocation logic. Agents developed using different frameworks can therefore communicate with user clients and other NLIP-compatible agents through a common protocol.

This extensibility has been demonstrated through the integration of NLIP into the AG2~\cite{ag2} agent development framework in April 2026~\cite{ag2_nlip_pr2468,ag2_nlip_pr3024}. AG2 reports more than one million monthly downloads (1,087,122 downloads in March 2026) and over 57,000 daily downloads with peaks exceeding 68,000, and is widely used in industry and research across diverse application domains. As AG2 adds support for NLIP, these deployments provide early evidence of NLIP's growing footprint as a standardized interaction layer for large-scale agent systems.

\subsection{Examples of NLIP Applications}

\noindent\textbf{Self-healing of Telecom equipment.} A commercial use case was demonstrated by IBM at the 2025 Telecommunications Management Forum (TMF)~\cite{TMForum}, featuring a collection of intelligent AI agents that would automatically diagnose the root cause of an equipment failure and (subject to applicable policies) automatically reboot or reconfigure the device at fault. The communication with the device was conducted using standards specified by TMF, with NLIP-based AI agents invoking those standards to monitor and reconfigure the devices. The system consisted of two Agents that leveraged the IBM Granite Mixture of Expert models to diagnose and fix the problem, one identifying the root cause (the Event Agent) and the other identifying the actions to be followed to inform the customer of any problem using a TMF Forum API. NLIP was used by the two intelligent agents to communicate with each other. 

As the system was being prepared for demonstration, a useful attribute of NLIP was discovered. In order to demonstrate the operations of the agents, a demonstration console was needed. While the typical demonstration console would have required supporting a special interface to get demonstration information from the different systems involved in this demonstration, NLIP provided a convenient interface to put a demonstration console above the system being developed. The existing component, namely the event collector, required a special interface to show what it was executing, but the NLIP-based agents could be demonstrated without requiring any additional protocol support.

\vspace{3pt}\noindent\textbf{Multimodal NLIP agent.} In this use case~\cite{6_ECMATR113}, three specialized agents, Speech to Text Agent, Channel Recommender Agent, and Search Agent, collaborate using NLIP to fulfill a customer’s voice-based support request in real time. Each agent plays a distinct role in the end-to-end interaction pipeline, showcasing how modular AI components can be composed using standardized communication and interoperable APIs.
The spoken audio is transcribed into text using a speech-to-text model. Based on keywords extracted from the customer’s request, the system identifies relevant Reddit channels. A specialized search agent then queries these channels and retrieves the most recent and relevant posts, delivering them back to the customer in real time. %

This transcribed content is then processed by a channel recommender agent, which may use an LLM enabling contextual understanding and intent extraction. Additionally, the search agent integrates MCP with Reddit’s search APIs to retrieve relevant, real-time information. This is an example of NLIP being used on the client- and agent-facing side for semantic interaction, while MCP is used internally to access tools and APIs. %
This use case illustrates how multi-agent systems can dynamically support customer queries across external knowledge sources.

\section{NLIP and Related Protocols}
\label{sec_other_protocol}

NLIP belongs to a broader family of agent communication efforts, from earlier agent communication languages such as the Knowledge Query and Manipulation Language (KQML)~\cite{7_KQML} and the Foundation for Intelligent Physical Agents Agent Communication Language (FIPA ACL)~\cite{8_FIPAACLMessageStructure,9_FIPACommunicativeActLibrary} to recent protocols such as the Model Context Protocol (MCP)~\cite{10_MCPSpecification}, Agent2Agent (A2A)~\cite{11_A2ASpecification}, the Agent Communication Protocol (ACP)~\cite{12_ACPDocumentation}, and the Agent Network Protocol (ANP)~\cite{13_ANPWhitePaper}. AG-UI additionally focuses on agent-to-user interaction, providing an event-based protocol for connecting agents with user-facing applications~\cite{27_AGUI}.  Other emerging efforts address adjacent parts of the agent ecosystem, including Media over QUIC Transport (MoQ)-based proposals for low-latency agent transport and trust-oriented protocols such as the Agent Trust Protocol (ATP) for agent identity, authorization, and attestation~\cite{24_MOQTransportForAgentProtocols,25_AgentProtocolOverMoQ,26_ATP}. Public NLIP project materials were available in 2024~\cite{21_NLIPProjectDocuments}, followed by an overview paper presented in March 2025 at the AAAI '25 Workshop on Open-Source AI for Mainstream Use~\cite{22_NLIPOverview}. \looseness=-1

NLIP is best understood as a neutral application-layer semantic interaction protocol that complements, rather than replaces, existing agent, tool, and enterprise protocols. In Mode A, an NLIP-aware agent or gateway exposes an NLIP-facing interface to clients and peer agents, while using protocols such as MCP, A2A, ACP, enterprise APIs, or industrial protocols internally or toward underlying services. In Mode B, NLIP-aware agents mediate interoperability among agents implemented with different underlying protocols. This allows NLIP to serve as a semantic coordination layer across a fragmented protocol ecosystem, while preserving existing investments in tool protocols, agent frameworks, and enterprise infrastructure.
Compared with narrower tool-access or agent-to-agent protocols, NLIP addresses a broader semantic translation boundary, making explicit grounding by NLIP-aware agents especially important for preserving domain-specific context.
\looseness=-1

\vspace{3pt}\noindent\textbf{An analogy to the Web.}
At the agent-ecosystem level, NLIP is designed to play a role analogous to HTTP/HTTPS on the Web: a common interaction layer that allows independently implemented endpoints to communicate through standardized messages.
Just as Web browsers and servers exchange HTTP/HTTPS messages regardless of the application semantics carried above them, AI agents can exchange NLIP messages regardless of their internal frameworks, task structures, workflow patterns, or execution environments. Higher-level semantic abstractions, such as A2A tasks and agent cards, can naturally be carried within NLIP messages, much like HTML documents, REST APIs, or application-specific data formats are transported over HTTP/HTTPS. This layering makes NLIP suitable as a general interoperability substrate, while allowing more specialized protocols to define higher-level workflow semantics above or alongside it.

\subsection{An Empirical Comparison of NLIP and A2A}
\label{sec_comparision_with_a2a}

\vspace{3pt}\noindent\textbf{Design goals.}
At a high level, both NLIP and A2A support agent-to-agent communication, but they make different design choices. A2A originated from Google and follows a task-oriented client-server model in which a client agent delegates work to a remote agent~\cite{Google_A2A_CollaborativeAgents}. Its protocol model includes predefined structures such as tasks, task states including ``input-required''~\cite{A2ATaskConcept}, and agent cards with fixed fields~\cite{A2AAgentCardConcept}. These abstractions support structured task delegation, but also impose a specific task model. In contrast, NLIP does not hard-code tasks, task states, workflows, or delegation relationships into its core message model; agents define interaction semantics through NLIP messages, enabling arbitrary conversations, workflows, and coordination patterns across heterogeneous systems.

\vspace{3pt}\noindent\textbf{Performance overhead.}
A2A's richer predefined structures can support governance-intensive workflows, but may add runtime overhead in connection handling, message exchange, serialization, and task-state management. In a controlled comparison using the same proof-of-concept three-agent customer-support workflow, NLIP achieved 8.4--9.6$\times$ lower latency than A2A-SDK on the lightweight coordination stage across two hardware environments (Apple M1 with 16 GB memory and Apple M2 Pro with 32 GB memory), narrowing to about 4$\times$ on a third, faster machine (Apple M3 Pro with 36 GB memory)~\cite{23_NLIPA2AEfficiencyComparison, sinha2026costofahop}. These measurements bound the protocol envelope only: LLM inference runs at the agent endpoints, outside the message-creation, connection, and send phases measured here. 
Message creation overhead was negligible, and A2A-SDK connection establishment dominated the gap. With connection caching enabled the advantage becomes hardware-dependent: NLIP retained a 2.75$\times$ lead on one machine but only 1.27$\times$ on a faster one. Against the more optimized Python-A2A, NLIP maintained a 4.1--4.3$\times$ advantage on the same stage~\cite{23_NLIPA2AEfficiencyComparison, sinha2026costofahop}. These results suggest that NLIP may be well suited for frequent, lightweight, latency-sensitive interactions, while A2A-style task abstractions are better suited for long-running or governance-intensive workflows and can be carried over, integrated with, or bridged through NLIP.

\section{Conclusion}
\label{sec_conclusion}

NLIP is a formal standard specifically designed for natural-language-centered interaction among modern AI agents, and broader adoption is emerging. NLIP's broader significance is that it standardizes the semantic envelope for interaction, while NLIP-aware agents and gateways provide the adaptive glue across heterogeneous, domain-specific agent ecosystems. This makes NLIP a candidate interoperability layer for multi-protocol, ontology-rich, and locally controlled AI deployments.

\section*{ACKNOWLEDGEMENT}
The authors from University of Illinois at Urbana-Champaign (UIUC) are supported in part by NSF TI-2625398.

\bibliographystyle{ACM-Reference-Format}
\bibliography{references}

\end{document}